\documentclass[10pt,twocolumn,letterpaper]{article}

\usepackage{cvpr}
\usepackage{times}
\usepackage{epsfig}
\usepackage{graphicx}
\usepackage{amsmath}
\usepackage{amssymb}

\usepackage[breaklinks=true,colorlinks,bookmarks=false]{hyperref}

\cvprfinalcopy 

\def\cvprPaperID{2313} 

\begin{document}

\title{NOTE-RCNN: NOise Tolerant Ensemble RCNN for Semi-Supervised Object Detection}

\author{Jiyang Gao$^{1}$ \quad Jiang Wang$^{2}$ \quad Shengyang Dai$^2$  \quad Li-Jia Li$^2$ \quad Ram Nevatia$^1$ \\
$^1$University of Southern California \qquad $^2$Google Could \\
{\tt\small \{jiyangga, nevatia\}@usc.edu,\quad \{wangjiang, sydai\}@google.com,\quad lijiali@cs.stanford.edu} 
}

\maketitle

\begin{abstract}
   The labeling cost of large number of bounding boxes is one of the main challenges for training modern object detectors. To reduce the dependence on expensive bounding box annotations, we propose a new semi-supervised object detection formulation, in which a few seed box level annotations and a large scale of image level annotations are used to train the detector. We adopt a training-mining framework, which is widely used in weakly supervised object detection tasks. However, the mining process inherently introduces various kinds of labelling noises: false negatives, false positives and inaccurate boundaries, which can be harmful for training the standard object detectors (\eg Faster RCNN). We propose a novel NOise Tolerant Ensemble RCNN (NOTE-RCNN) object detector to handle such noisy labels. Comparing to standard Faster RCNN, it contains three highlights: an ensemble of two classification heads and a distillation head to avoid overfitting on noisy labels and improve the mining precision, masking the negative sample loss in box predictor to avoid the harm of false negative labels, and training box regression head only on seed annotations to eliminate the harm from inaccurate boundaries of mined bounding boxes. We evaluate the methods on ILSVRC 2013 and MSCOCO 2017 dataset; we observe that the detection accuracy consistently improves as we iterate between mining and training steps, and state-of-the-art performance is achieved.
\end{abstract}

%

\section{Introduction}

With the recent advances in deep learning, modern object detectors, such as Faster RCNN \cite{ren2015faster}, YOLO \cite{redmon2016you}, SSD \cite{liu2016ssd} and RetinaNet \cite{lin2017focal}, are reliable in predicting both object classes and their bounding boxes. However, the application of deep learning-based detectors is still limited by the efforts of collecting bounding box training data. These detectors are trained with huge amount of manually labelled bounding boxes. In real world, each application may require us to detect a unique set of the categories. It's expensive and time-consuming to label tens of thousands of object bounding boxes for each application. 

\begin{figure}[]
  \centering
    \includegraphics[width=0.45\textwidth]{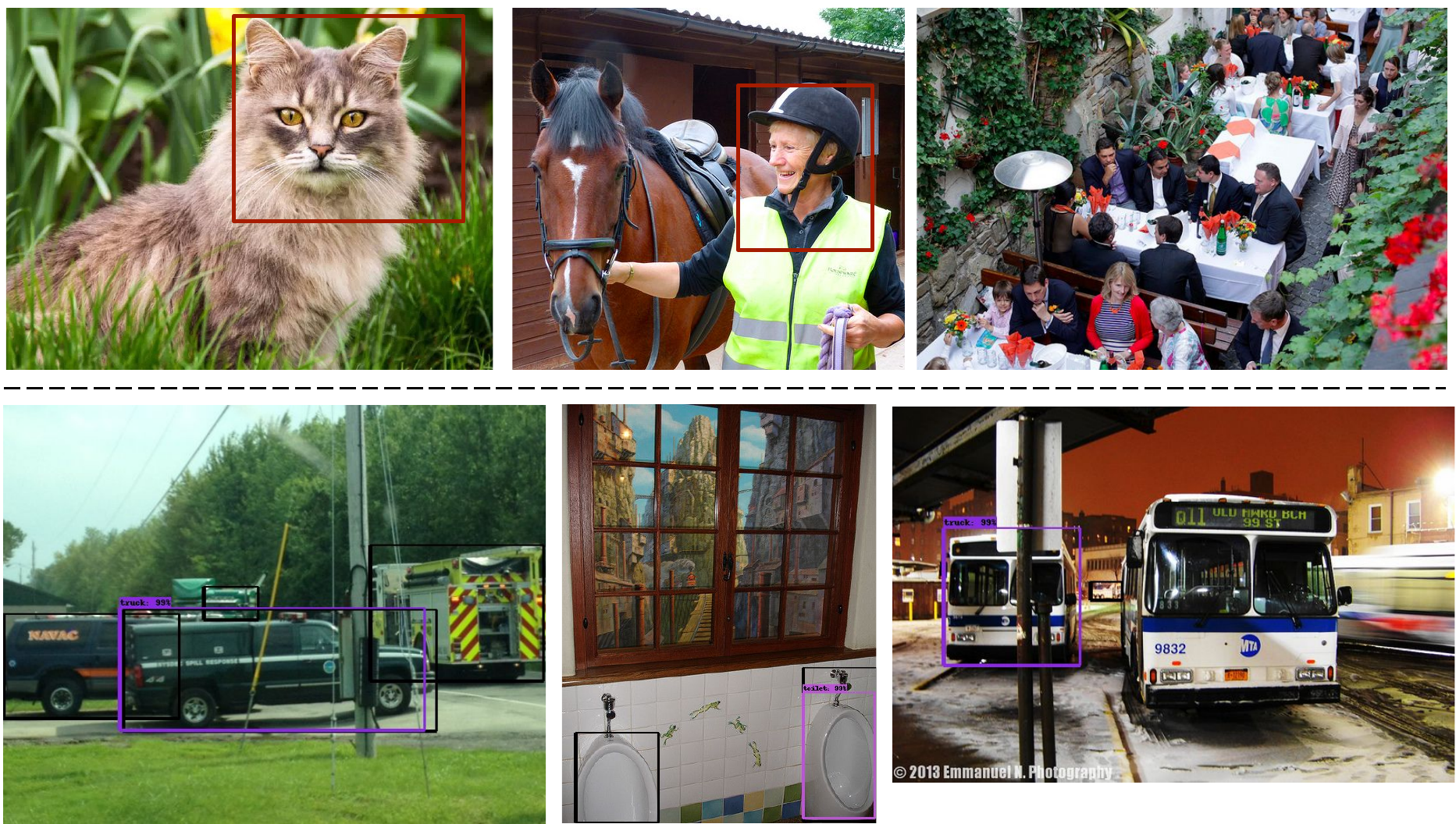}
    \caption{Top: examples of weakly supervised object detection failure cases: poor localization; objects can't be discovered in complicated scenes. Bottom: examples of the mined box noises using a standard faster RCNN: 1) false negatives, 2) false positives, 3) inaccurate box boundaries; groundtruth boxes are in black, mined boxes are in other colors. }
      \label{fig:baseline-example}
\end{figure}

To reduce the effort of labelling bounding boxes, researchers worked on learning object detectors with only image-level labels, which are substantially cheaper to annotate, or even free with image search engines; this task is called weakly supervised object detection \cite{cinbis2017weakly,tang2017multiple,Zhang_2018_CVPR}. Multiple Instance Learning (MIL) \cite{dietterich1997solving} based training-mining pipeline \cite{cinbis2017weakly, tang2017multiple, Uijlings_2018_CVPR} is widely used for this task; however, the resulting detectors perform considerably worse than the fully supervised counterparts. We believe the reasons are two-fold: First, a detector learned with only image-level labels often performs poorly in localization, it may focus on the object part, but not the whole object (\eg, in Figure \ref{fig:baseline-example}, ``cat" detector detects cat head); second, without an accurate detector, object instances cannot be mined correctly, especially when the scene is complicated.

\begin{figure}[]
  \centering
    \includegraphics[width=0.43\textwidth]{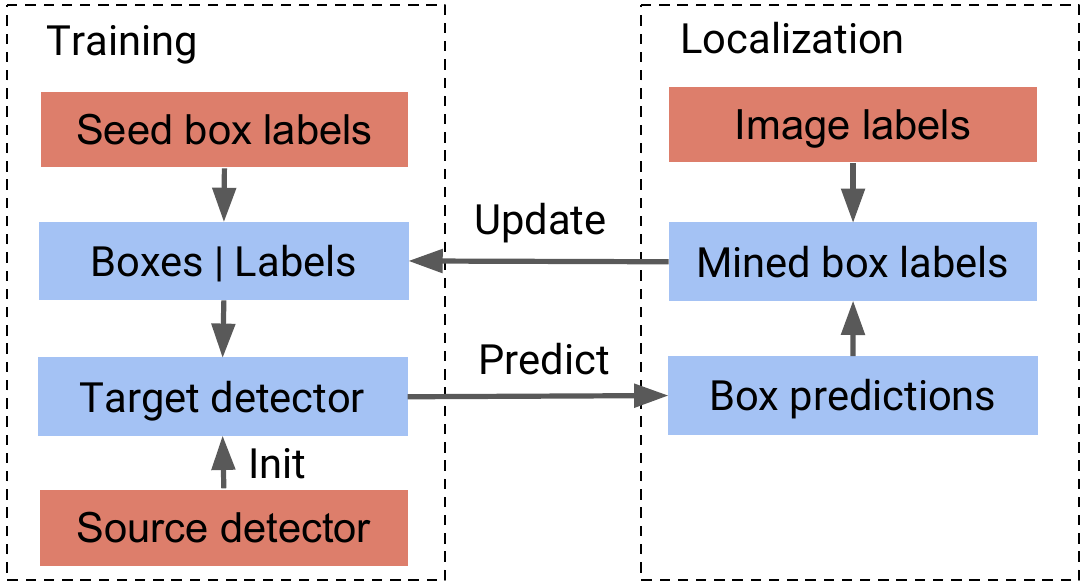}
    \caption{Iterative training-mining pipeline.}
      \label{fig:train-localize}
\end{figure}

To address the aforementioned problems in weakly supervised object detection, we propose a semi-supervised object detection setting: learning an object detector with a limited amount of labelled bounding boxes (e.g. 10 to 20 images with fully labeled bounding boxes) as well as a large amount of image-level labels. Specifically, we want to train an object detector for a set of \emph{target} categories. For target categories, a small amount of \emph{seed} bounding box annotations and a large amount of image-level annotations are available for training. We also assume that a pre-trained object detector for \emph{source} categories is available. The source and target categories do not overlap with each other. Given the wide availability of large scale object detection datasets, such as MSCOCO \cite{lin2014microsoft} and ILSVRC  \cite{russakovsky2015imagenet}, this assumption is not hard to satisfy in practice. This assumption is not essential for the formulation either.
Note that our formulation is different from previous semi-supervised object detection  \cite{hoffman2014lsda,tang2016large}, in which the seed bounding box annotations are not considered. 

The standard training-mining pipeline \cite{cinbis2017weakly,tang2017multiple} in weakly supervised object detection iterates between the following steps: 1. Train object detector with the mined bounding boxes (the initial detector is trained with the whole images and the labels); 2. Mine the bounding boxes with the current object detector. A straight-forward way to incorporate the seed bounding boxes is that we use them to train the initial object detector, mine bounding boxes with the initial object detector, train a new detector with both seed and mined bounding boxes, and iterate between mining and training steps. 

The mining process inherently introduces various types of noise. First, mining process inevitably misses some objects, which are treated as negative (\ie background) samples in training phase; such false negatives are harmful for training the classification head of object detector. Second, the boundaries of the mined bounding boxes are not precise, which is harmful for learning the box regression head of the detector. Third, the class labels of the mined boxes cannot be 100\% accurate, leading to some false positives. Some visualization examples for the mined labels from the baseline method are shown in Figure \ref{fig:baseline-example}. Because of these issues, we observe that the detection accuracy usually decreases as we iterate between training and mining steps if standard object detector architecture (\eg Faster RCNN) is employed.

We propose a novel NOise Tolerant Ensemble RCNN (NOTE-RCNN) architecture. The NOTE-RCNN incorporates an ensemble of classification heads for both box predictor (\ie second stage) and region proposal predictor (\ie first stage) to increase the precision of the mined bounding boxes, \ie, reduce false positives. Specifically, one classification head is only trained with seed bounding box annotations; the other head is trained with both seed and mined box annotations. The consensus of the both heads is employed to determine the confidence of the classification. This is similar to recent work that uses ensemble for robust estimation of prediction confidence~\cite{choi2018generative, buckman2018sample}. 
We also utilize the knowledge of the pre-trained detector on source categories as \emph{weak teachers}. Specifically, another classification head is added to distill knowledge \cite{hinton2015distilling} from a weak teacher; the distillation process acts as a regularizer to prevent the network from overfitting on the noisy annotations. 
The NOTE-RCNN architecture is also designed to be robust to false negative labels. 
For the classification head in the box predictor that uses mined bounding boxes for training, we remove the loss of predicting negatives (\ie background) from its training loss, thus the training is not affected by the false negatives.
Finally, the regression head is only trained with the seed bounding boxes, which avoids it being affected by the inaccurate boundaries of the mined bounding boxes.

We evaluated the proposed architecture on MSCOCO \cite{lin2014microsoft} and ILSVRC \cite{russakovsky2015imagenet} datasets. The experimental results show that the proposed framework increases the precision of mined box annotations and can bring up to 40\% improvement on detection performance by iterative training. Compared with weakly supervised detection, training with seed annotations using NOTE-RCNN improves the state-of-the-art performance from 36.9\% to 43.7\%, while using standard Faster RCNN only achieves 38.7\%.
We also perform a large scale experiment which employs MSCOCO as seed annotations and 
Open Image Dataset as image-level annotations. 
We observe the proposed method also leads to consistent performance improvement 
during the training-mining process.

In summary, our contributions are three-fold: first, we propose a practical semi-supervised object detection problem, with a limited amount of labelled bounding boxes as well as a large amount of image-level labels; second, we identified three detrimental types of noise that inherently exists in training-mining framework ; third, we propose a novel NOTE-RCNN architecture that is robust to such noise, and achieves state-of-the-art performance on benchmark datasets.

\section{Related Work}
\textbf{Weakly supervised object detection.}
The majority of recent work treats weakly supervised object detection as a Multiple Instance Learning (MIL) \cite{dietterich1997solving} problem. An image is decomposed into object proposals using proposal generators, such as EdgeBox \cite{zitnick2014edge} or SelectiveSearch \cite{uijlings2013selective}. The basic pipeline is to iteratively mine (\ie localize) objects as training samples using the detectors and then train detectors with the updated training samples. The detector can be a proposal level SVM classifier \cite{song2014learning, cinbis2017weakly, Uijlings_2018_CVPR} or modern CNN-based detector \cite{tang2017multiple, jie2017deep, Zhang_2018_CVPR}, such as RCNN \cite{girshick2014rich} or Fast RCNN \cite{girshick2015fast}. Deselaers \etal \cite{deselaers2012weakly} first argued to use objectness score as a generic object appearance prior to the particular target categories. Cinbis \etal \cite{cinbis2017weakly} proposed a multi-fold multiple instance learning procedure, which prevents training from prematurely locking onto erroneous object locations. Uijlings \etal \cite{Uijlings_2018_CVPR} argued to use pre-trained detectors as the proposal generator and show its effectiveness in knowledge transfer from source to target categories.

Recently, there are also work that designs end-to-end deep networks combining with multiple instance learning. Bilen \etal \cite{bilen2016weakly} designed a two-stream network, one for classification and the other for localization, it outputs final scores for the proposals by the element-wise multiplication on the scores from the two streams. Kantorov et al. \etal \cite{kantorov2016contextlocnet} proposed a context-aware CNN model
based on contrast and additive contextual guidance, which improved the object localization accuracy.

\textbf{Semi-supervised object detection.}
Note that in previous work \cite{hoffman2014lsda}, the definition of semi-supervised object detection is slightly different from ours, in which only the image-level labels and pre-trained source detectors are considered, but seed bounding box annotations are not used.
Beginning from LSDA \cite{hoffman2014lsda}, Hoffman \etal proposed to learn parameter transferring functions between the classification network and the detection network, so that a classification model trained by image level labels can be transferred to a detection model. Tang \etal \cite{tang2016large} explored the usage of visual and semantic similarities among the source categories and the target categories in the parameter transferring function. Hu \etal \cite{Hu_2018_CVPR} further extended this method to semi-supervised instance segmentation, which transfers models for object detection to instance segmentation. Uijlings \etal \cite{Uijlings_2018_CVPR} adopted the MIL framework from weakly supervised object detection, and replaced the unsupervised proposal generator \cite{zitnick2014edge} by the pre-trained source detectors to use the shared knowledge. Li \etal \cite{Li_2018_CVPR} proposed to use a small amount of location annotations to simultaneously performs disease identification and localization.

\section{Method}
We first briefly introduces our semi-supervised object detection learning formulation and training-mining framework. We present the proposed NOise Tolerant Ensemble R-CNN (NOTE-RCNN) detector. 

\subsection{Problem Formulation}
We aim to train object detectors for target categories. 
For target categories, we have a small amount of seed bounding box annotations $\mathbf{B^0}$, 
as well as a large amount of image level annotations $\mathbf{A}$.
We also have a pre-trained object detection $\mathcal{S}$ on source categories, which do not overlap with target categories.

\subsection{Detector Training-Mining Framework}
The object detectors are trained in a iterative training-mining framework, where the trained detector at iteration $t$ is denoted as $\mathcal{T}^t$.
The detector training-mining framework has the following steps.

\textbf{Detector Initialization.} 
A initial target detector $\mathcal{T}^0$ is initialized from the source detector $\mathcal{S}$ 
and trained using the seed bounding box annotations $\mathbf{B^0}$. 

\textbf{Box Mining.} 
Box mining uses the the current detector $\mathcal{T}^{t-1}$ to mine a set of high quality bounding box annotation 
$\mathbf{B^t}$ for target categories from annotations with image-level labels $\mathbf{A}$.
A bounding box is mined if it fulfills the following conditions: 
1) its (predicted) label matches with the image-level groundtruth label;
2) the box's confidence score is the highest among all boxes with the same label;
3) its confidence score is higher than a threshold $\theta_b$.
The process can be summarized as $\mathbf{B^t}=\mathbf{M}(\mathbf{A},\mathcal{T}^{t-1},\theta_b)$, where $\textbf{M}$ is the box mining function; 

\textbf{Detector Retraining.}  
A new detector $\mathcal{T}^{t}$ is trained with the union of mined bounding boxes $\mathbf{B^t}$ and the seed bounding boxes $\mathbf{B^0}$. 
The parameters of the new detector $\mathcal{T}^t$ are initialized from the detector $\mathcal{T}^{t-1}$ from the previous iteration. 
The process can be summarized as $\mathcal{T}^{t}=\mathbf{R}(\mathbf{B^t},\mathbf{B^0},\mathcal{T}^{t-1})$, where $\mathbf{R}$ represents the re-training function.

\begin{figure*}[]
  \centering
    \includegraphics[width=0.9\textwidth]{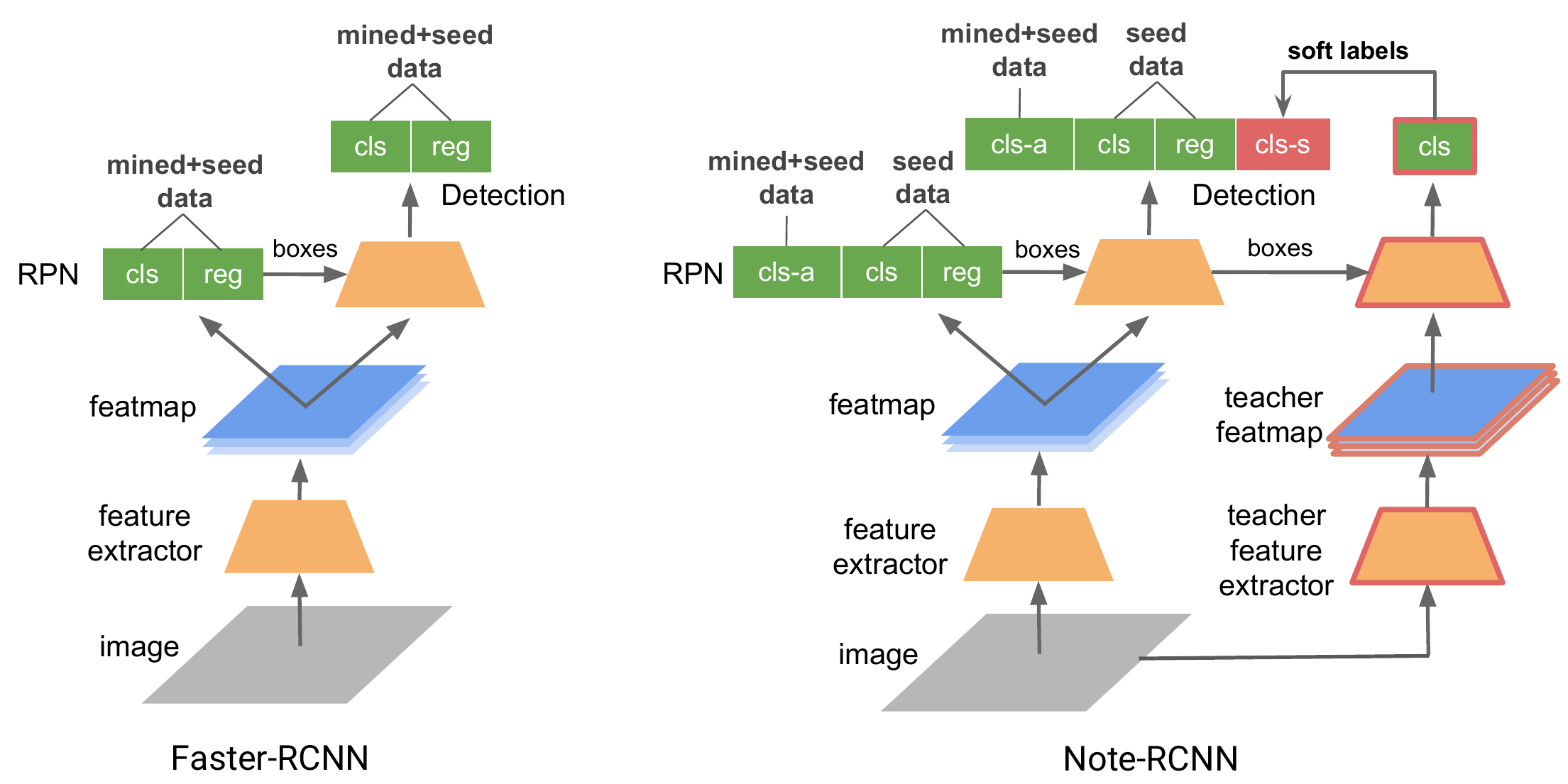}
    \caption{NOise Tolerant Ensemble R-CNN (NOTE-RCNN) architecture. There are three differences from standard Faster RCNN: additional classification heads in RPN and box predictor; noise tolerant training strategy; knowledge distillation from pre-trained detectors.}
      \label{fig:ss-rcnn-model}
\end{figure*}

\subsection{NOTE-RCNN}
NOTE-RCNN is designed to be tolerate to noisy box annotations that are generated in the training-mining framework.

There are three types of noise in the mined boxes: false negatives, false positives and box coordinate noise of the mined boxes. 
NOTE-RCNN is based on Faster RCNN, and it contains three improvements for noise tolerance: 
ensembling two classification heads and a distillation head to avoid overfitting on noisy labels and improve the mining precision, 
masking the negative sample loss in box predictor to get rid of the harm of false negative labels, 
and training box regression head only on seed annotations to eliminate the effect of inaccurate box coordinates of mined bounding boxes.

\subsubsection{Recap of Faster RCNN architecture}
In Faster RCNN, object locations are predicted in two stages: proposal prediction stage and box prediction stage. 
The first stage, called Region Proposal Network (RPN), outputs a set of class-agnostic proposal boxes for an image. 
It uses a feature extractor (\eg, VGG-16, ResNet-101) to extract feature maps from an image, 
and it predicts proposal boxes using ROI pooled features in a set of predefined anchors in this feature map.
We denote its classification head as \textbf{rpn-cls}, the box coordinate regression head as \textbf{rpn-reg}. 
An outline of the architecture is shown in the left part of Figure \ref{fig:ss-rcnn-model}. 
The loss function of RPN is as follows:
\begin{equation}
\label{loss}
\begin{aligned}
    L(\{p_i\},\{t_i\})=\frac{1}{N_{cls}}\sum_i L_{cls}(p_i, p_i^*)+\\
    \frac{1}{N_{reg}}\sum_i p_i^* L_{reg}(t_i, t_i^*)
\end{aligned}
\end{equation}
where $i$ is the index of an anchor and $p_i$ is the predicted object probability of anchor $i$. 
The ground-truth label $p_i^*$ is $1$ if the anchor's overlap with groundtruth bounding box is larger than a threshold, and is 0 otherwise.
$t_i$ is a vector encoding of the coordinates the bounding box, and $t_i^*$ is that of the ground-truth box associated with a positive anchor, 
$L_{cls}=-p_i^*log(p_i)$ is binary cross-entropy loss, 
$L_{reg}$ is smooth L1 loss.

In the second stage, called box predictor network, 
Features are cropped from the same intermediate feature maps for each proposal box, and they are resized to a fixed size.
These features are fed to the box predictor network to predict class probabilities and class-specific box refinement for each proposal. 
We denote the classification head as \textbf{det-cls}, 
the boundary regression head as \textbf{det-reg}. 
The loss function for the second stage is similar to Equation \ref{loss}. 
The only difference is that $p_i$ is replaced by $p_i^u$, which is the predicted probability of category $u$; correspondingly, $p_i^{u*}$ is 1 if the proposal box's 
overlap with a box with category $u$ is larger than a threshold.

\subsubsection{Box Predictor in NOTE-RCNN}
In order to improve the noise tolerance, we use an ensemble of two classification heads in box predictor network (\ie second stage of the detector). 
The seed classification head \textbf{det-cls} is trained only one seed bounding box annotations $\mathbf{B^0}$ so that it 
is not disturbed by the false negatives and false positives in the mined annotations $\mathbf{B^t}$. The mixed classification head \textbf{det-cls-a}
utilizes both seed box annotations $\mathbf{B^0}$ and mined box annotations $\mathbf{B^t}$ for training. 
The consensus of the seed and mixed classification head is employed for a robust estimation of classification confidence.

The regression head \textbf{det-reg} is trained only one seed bounding box annotations $\mathbf{B^0}$, too, so that it is  not affected by the inaccurate box coordinates in $\mathbf{B^t}$.

\textbf{Filtering background proposal loss.} 
Given that false negative is extremely hard to eliminate in mined bounding boxes, 
the losses of ``background" proposals in \textbf{det-cls-a} are not used in loss to 
remove the effect of false negatives.

 Specifically, if an image $i$ is from mined box annotation set $\mathbf{B^t}$, 
 then we mask the losses from the proposals that belong to ``background" category (typical implementation uses index 0 for background); 
 if the image is from seed box annotation set $\mathbf{B^0}$, then the loss is calculated normally. 
 The classification loss can be expressed as
\begin{equation}
\begin{aligned}
    L_{det-cls-a}(p_i,u,i)=-p^{u*}_ilog(p_i^u)*\lambda(u,i),\\
    \lambda(u,i)=
\begin{cases}
0& \text{u=0} ~\& ~i\notin \mathbf{B^0}\\
1& \text{otherwise}
\end{cases}
\end{aligned}
\end{equation}

During training, the loss function for the box predictor consists of the losses ofthree heads: \textbf{det-cls}, \textbf{det-reg} and \textbf{det-cls-a}, i.e., $L_{det}=L_{det-cls}+L_{det-cls-a}+L_{det-reg}$. 
During inference, the classification probability outputs from \textbf{det-cls} and \textbf{det-cls-a} are averaged.

\subsubsection{RPN in NOTE-RCNN}
Similarly, we add an additional binary classification head \textbf{rpn-cls-a} in RPN. 
Similar to box predictor, the seed classification head \textbf{rpn-cls} and the regression head \textbf{rpn-reg} are trained only on seed bounding box annotations $\mathbf{B^0}$. 
The mixed head \textbf{rpn-cls-a} uses both seed box annotations $\mathbf{B^0}$ and mined box annotations $\mathbf{B^t}$ for training. 
Different from box predictor, we don't zero the background loss if the training image is from mined annotation set, as RPN solves a binary classification problem and filtering background loss makes it unlearnable.

During training, the loss function for RPN comes from the three heads, \textbf{rpn-cls}, \textbf{rpn-reg} and \textbf{rpn-cls-a}, which can be expressed as $L_{rpn}=L_{rpn-cls}+L_{rpn-cls-a}+L_{rpn-reg}$. 
During inference, the classification probability outputs from \textbf{rpn-cls} and \textbf{rpn-cls-a} are averaged.

\subsubsection{Knowledge Distillation as Supervision}
We added a knowledge distillation head \textbf{det-cls-s} to source detector $S$ for additional noise tolerance, because it stops the target detector from overfitting to noisy annotations.
During training, for a image $I_k$, we first forward $I_k$ in the target detectors $\mathbf{T^t}$ to generate proposal boxes $\{P^t_k\}$.
Then we forward the image $I_k$ together with the proposals $P^t_k$ to the source detector $S$ to get the probability distribution on the source classes for every proposal. 
We use such distribution as a supervision to train \textbf{det-cls-s}. 
This process is known as knowledge distillation \cite{hinton2015distilling}. The loss function can be expressed as 
\begin{equation}
   L_{dist}=\frac{1}{N_{dist}}\sum_s \sum_j -p_s^{j*}log(p_s^j) 
\end{equation}
where $j$ is the class index, $s$ is the proposal index, $p_s^{j*}$ is the probability of proposal $s$ for class $j$ from source detectors, and $p_s^{j}$ is that from target detectors. 
The gradients generated from \textbf{det-cls-s} don't affect the parameters of \textbf{det-cls-a}, \textbf{det-cls} and \textbf{det-reg}, but they affect the feature extractor parameters. 

As the source detectors are trained on large scale clean annotations, we expect to use probability distribution generated from source detectors as additional supervision to regularize the feature extractor in target detectors. Our motivation is not to directly affect the classification head of target categories, but to prevent the feature extractor from overfitting the noisy annotations.



\section{Evaluation}
In this section, we first present the implementation details of the whole detection system. 
We then introduce the benchmark datasets for evaluation. 
Third, we introduce our evaluation metrics and ablation studies. 
Finally, we discuss the experimental results on MSCOCO and ILSVRC, as well as Open Image Dataset (OID).

\subsection{Implementation Details}
We use Inception-Resnet-V2~\cite{szegedy2017inception} as the feature extractor of the detector for all the experiments in this paper. 
The Inception-Resnet-V2 feature extractor is initialized from
the weights trained on ImageNet classification dataset \cite{russakovsky2015imagenet}.
All input images are resized to $600*1024$. 
In the first stage, 300 proposal boxes are selected. 
We use SGD with momentum with batch sizes of 1 and learning rate at 0.0003. 
The system is implemented using the Tensorflow Object Detection API \cite{huang2017speed}. 
In all the experiments except the OID one, 
we employ 8 iterations of training-mining process,
because we find the performance generally satuates after 8 iterations.
In each iteration, the model is trained for 20 epochs. 
The mining threshold $\theta_b$ is set to 0.99 if no other
specification is given.

\subsection{Datasets}
\textbf{MSCOCO 2017.} 
MSCOCO \cite{lin2014microsoft} contains 80 categories, which is a superset of PASCAL VOC \cite{everingham2015pascal} categories. 
We split both training and validation data to VOC categories (\ie source categories) and non-VOC categories (\ie target categories).
If an image has both source category and target category bounding boxes, this image is used in both source category data and target category data, 
but source category and target category data only contains bounding boxes with the corresponding categories.

The source category training data is used to train source detectors.
For target category training data, we randomly pick certain amount of images for each category as seed groundtruth bounding box annotations, and keep only image-level 
labels for the rest of images.
We evaluate the target detectors on the target category validation data. 
To evaluate the method under varied amounts of seed annotations, we experiment with 
seed annotations with different average sizes: [12, 33, 55, 76, 96].
The average size means the average number of annotated images per category.


\begin{figure}[]
  \centering
    \includegraphics[width=0.45\textwidth]{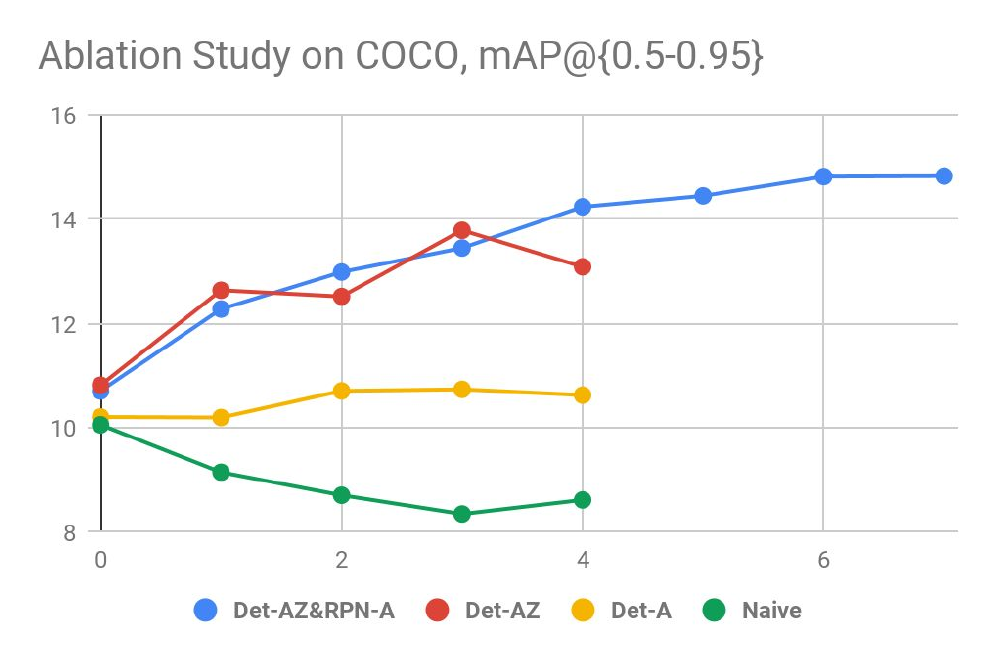}
    \caption{Ablation studies on MSCOCO 2017 dataset.}
      \label{fig:ablation-coco}
\end{figure}

\textbf{ILSVRC 2013.}  
We follow the same settings as \cite{hoffman2014lsda,tang2016large,Uijlings_2018_CVPR} for direct comparisons on ILSVRC 2013 \cite{russakovsky2015imagenet}. 
We split the ILSVRC 2013 validation set into two subsets val1 and val2, and augment val1 with images from the ILSVRC 2013 training set such that each class has 1000 annotated bounding-boxes in total \cite{girshick2014rich}. 
Among the 200 object categories in ILSVRC 2013, we use the first 100
in alphabetical order as sources categories and rest as target categories. 
We use all images of the source categories in augmented val1 set as the source training set, and that of the target categories in val2 set as source validation set.
For target training set, we randomly select 10-20 images for each target category in augment val1 set as seed groundtruth bounding boxes, and use the rest of images as
image-level labels by removing the bounding box information. 
All images of the target categories in val2 set are used as target validation set.

\textbf{OpenImage v4.} The training set of OpenImage V4 \cite{OpenImages} contains 14.6M bounding boxes for 600 object classes on 1.74M images. 
We only keep the image-level labels for those images by removing all bounding box 
information.
We use the whole MSCOCO dataset as the seed bounding box annotation set $\mathbf{B^0}$, and the aforementioned OpenImage images as image-level annotations $\mathbf{A}$ to evaluate whether our method can improve performance when we already have a large
number of groundtruth bounding boxes.

\subsection{Experiment Settings}
\label{exp-setting}
\textbf{Evaluation metric.}
For object detection performance, we use the mean Average Precision (mAP), which is averaged mAP over IOU thresholds in [0.5 : 0.05 : 0.95].
We also report mAP@IOU 0.5. 
To measure the quality of mined box annotations, we report \textit{box recall} and \textit{box precision}. 
Box recall means the percentage of the true positive boxes in the mined annotations over all groundtruth boxes, 
Box precision means the percentage of the true positive boxes over all boxes in the mined annotations.


\begin{figure}[]
  \centering
    \includegraphics[width=0.45\textwidth]{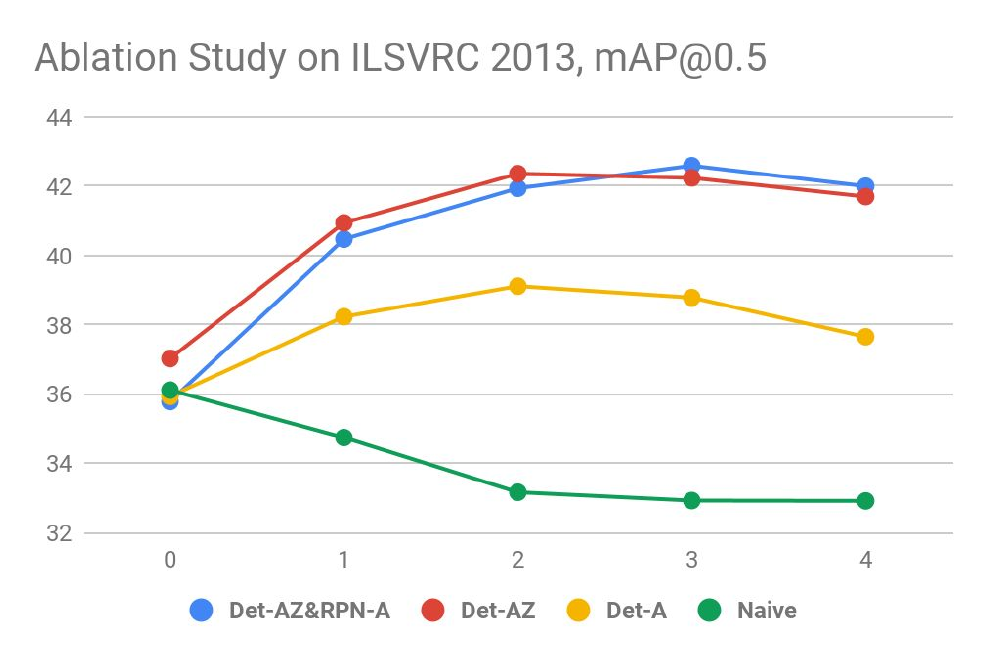}
    \caption{Ablation studies on ILSVRC 2013 dataset.}
      \label{fig:ablation-ilsvrc}
\end{figure}

\textbf{Ablation studies.}
To evaluate the contribution of each component in NOTE-RCNN, we design the following system variants for ablation studies: 

(1) \textbf{Naive}: no additional classification head is added to RPN nor box predictor, \ie stardard Faster RCNN; 
both mined annotation and seed groundtruth annotation are used to train the classification heads and regression heads (for both detection and RPN). 
(2) \textbf{Det-A}: we add the additional classification head \textbf{det-cls-a} to box predictor, but not to RPN; the original head \textbf{det-cls} and \textbf{det-reg} are trained by the seed groundtruth annotations, \textbf{det-cls-a} is trained on both seed groundtruth data and seed annotation; we don't zero the background sample loss in this variant. 
(3) \textbf{Det-AZ}: Similar to Det-A, but we zero the background sample loss in this variant. 
(4) \textbf{Det-AZ\&RPN-A}: we add the additional classification heads to both RPN and detection part. \textbf{det-cls}, \textbf{det-reg}, \textbf{rpn-cls}, \textbf{rpn-reg} are trained on the seed groundtruth annotations, \textbf{det-cls-a} and \textbf{rpn-cls-a} are trained on both seed annotations and mined annotations; We zero the background sample loss on \textbf{det-cls-a}, but not on \textbf{rpn-cls-a}. 
(5) \textbf{Det-AZ\&RPN-A\&Distill}: Similar to Det-AZ\&RPN-A, but we ad the distillation head.

\subsection{Experiments and Discussions}
\textbf{Evaluation on additional classification heads.} To show the contribution of each component, we do ablation studies on the additional heads and the system variants.
 Experimental results on MSCOCO are shown in Figure \ref{fig:ablation-coco}. 
 For Naive, Det-A and Det-AZ, we stop training in 4 iterations, as the performance  already decreases in iteration 4. 
 For Det-AZ\&RPN-A, we train it for 8 iterations. 
Experimental results on ILSVRC 2013 are shown in Figure \ref{fig:ablation-ilsvrc}.
On this dataset, We train all system variants for 4 iterations. 
Iteration 0 means that the detector is only trained on the seed groundtruth box annotations. The performances on iteration 0 for different system variants is slightly different, because we initialize each detector independently. 

From Naive models, we can see that if we don't separate the seed groundtruth annotations with mined annotations, and just train the regression head and classification head with all data, the performance drops immediately after we add the mined data (iteration 1 and after). 
For Det-AZ and Det-A, it can be observed that zeroing the background  loss gives significant improvements on both MSCOCO (in Figure \ref{fig:ablation-coco}) and ILSVRC (in Figure \ref{fig:ablation-ilsvrc}). 
Comparing Det-AZ\&RPN-A and Det-AZ in MSCOCO (Figure \ref{fig:ablation-coco}), we can see that the performance of Det-AZ\&RPN-A consistently increases, but that of Det-AZ starts to decrease after the 3rd iteration. 
We believe that the reason is that more accurate RPN and detection helps to improve the performance of each other. 
Specifically, the ensemble classification heads in RPN improve the proposal quality, resulting in the discovery of more object proposals; higher quality object proposals are beneficial to the detection performance; better detection performance leads to higher quality mined annotations, which in turn improves the RPN accuracy.
Thus, applying ensemble classification heads to both RPN and box predictor are important for consistent performance increase. The difference between Det-AZ\&RPN-A and Det-AZ on ILSVRC (in Figure \ref{fig:ablation-ilsvrc}) is not significant.
The reason is that ILSVRC 2013 is a relatively simple dataset for detection, where an image usually only contains 1 to 2 objects of interest and the area of object is usually large, leading to lower mining annotation difficulty.

\begin{figure}[]
  \centering
    \includegraphics[width=0.45\textwidth]{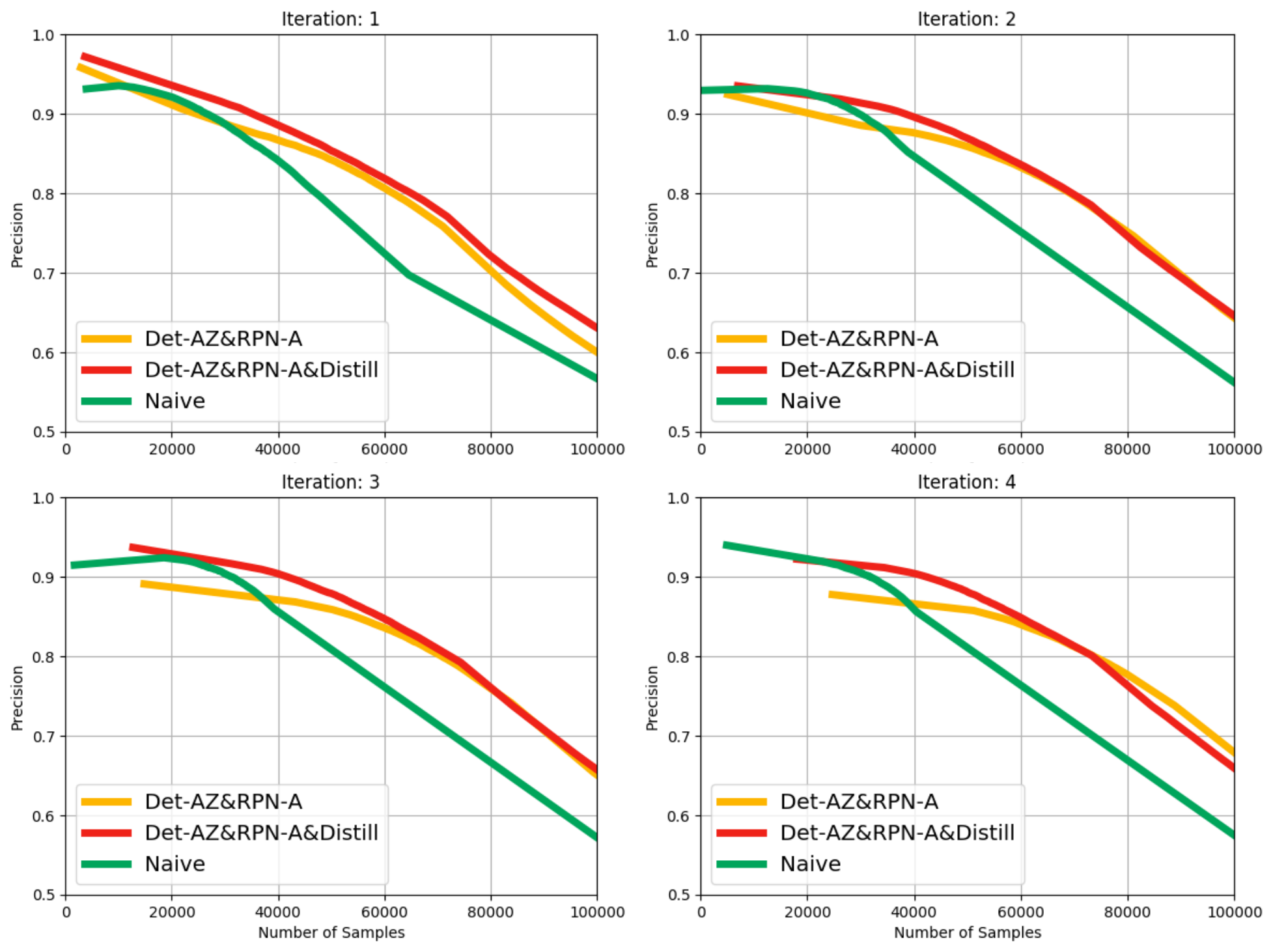}
    \caption{Comparison on ``Box Precision vs Number of Samples Curve" of Mined Annotations on MSCOCO 2017.}
      \label{fig:precision-number}
\end{figure}

\begin{figure}[]
  \centering
    \includegraphics[width=0.40\textwidth]{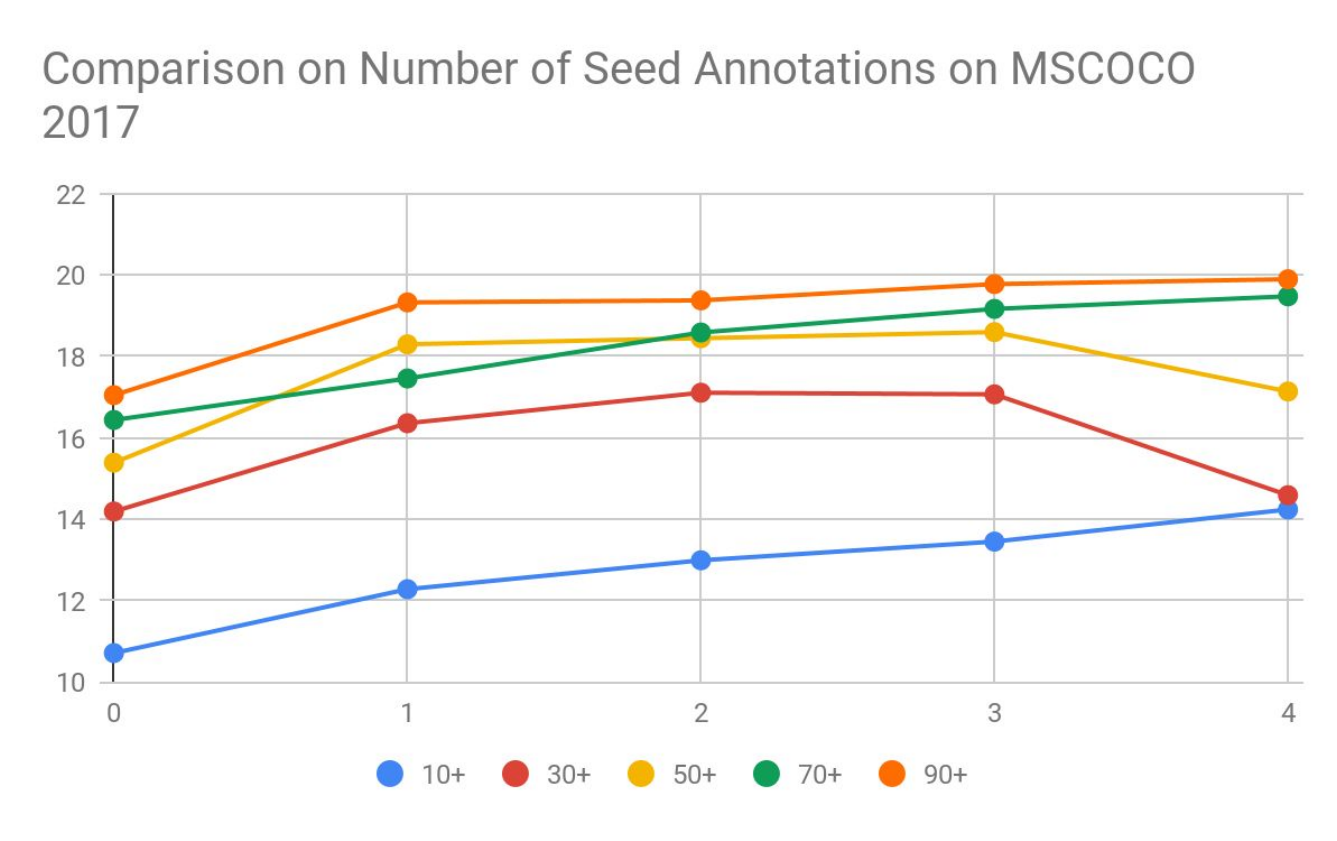}
    \caption{Comparison on different amount of seed annotations on MSCOCO 2017.}
      \label{fig:diff-seed}
\end{figure}

\textbf{Using different amount of seed annotations.} 
To evaluate the performance of the proposed method using different amount of seed bounding box annotations, we test NOTE-RCNN with varied sizes of seed annotation set on MSCOCO. 
The average sizes (\ie average number of annotated images per category) tested are [12, 33, 55, 76, 96]. 
The method used for evaluation is Det-AZ\&RPN-A. 
We can see in Figure \ref{fig:diff-seed} that NOTE-RCNN provides steady performance improvements for all experiments, indicating the effectiveness of the proposed method 
when different amount of seed annotated images are used. 

\begin{table}[]\small

\centering
\begin{tabular}{l|c|c|c|c}
\hline
\multicolumn{1}{c|}{} & \multicolumn{2}{c|}{Det-AZ\&RPN-A} & \multicolumn{2}{c}{Det-AZ\&RPN-A\&Distill} \\ \hline
\# iter   & \# boxes   & prec(\%)   & \# boxes    & prec(\%)         \\ \hline
1    &  21542    &  90.0  &   22972   &   88.3   \\ \hline
2    &  38323  &   87.1    &    32698   &   90.8    \\ \hline
3    &   44223  &     86.6  &   38727   &  89.9  \\ \hline
4    &   54680  &    84.9  &     41576 &     90.0   \\ \hline
5    &     60899 &  83.7   &      42756    &     89.9 \\ \hline
\end{tabular}
\vspace{0.5em}
\\\caption{Comparison between ``with distillation" and ``without distillation" on annotation mining on MSCOCO 2017, threshold $\theta_b$ is set to be 0.99.}
\label{tbl:distill}
\end{table}

\textbf{Bounding box mining quality.} 
We evaluate the bounding box mining precision for Naive, Det-AZ\&RPN-A and Det-AZ\&RPN-A\&Distill methods. 
First, we draw ``box precision vs number of samples" curves of mined annotations on MSCOCO, shown in Figure \ref{fig:precision-number}. 
This curve is generated by varying the mining threshold $\theta_b$ from 0 to 1.0, and we show the part of curve that falls in between $[0,10^5]$ samples.
The results of 1st to 4th iterations are shown. 
We can see that the precision of Naive drops very fast when the number of samples increase; Det-AZ\&RPN-A performs better than Naive when the number of samples is large; Det-AZ\&RPN-A\&Distill achieves the best precision performance. 

We further compare the actual precision and number of boxes in each iteration between Det-AZ\&RPN-A and Det-AZ\&RPN-A\&Distill by setting the $\theta_b$ as 0.99. As shown in Table \ref{tbl:distill}, we can see that: 
(1) without using distillation, the precision decreases gradually, from 90.0\% to 83.7\%, with distillation, the precision is preserved at around 90.0\%; 
(2) the increasing speed of mined box number of Det-AZ\&RPN-A is higher than that of Det-AZ\&RPN-A\&Distill. 
Generally, it can be seen that Det-AZ\&RPN-A performs better than Naive, which shows the effectiveness of the ensemble classification heads, and using distillation head further improves the mining precision by preventing the network from overfitting noisy labels.

\begin{figure}[]
  \centering
    \includegraphics[width=0.45\textwidth]{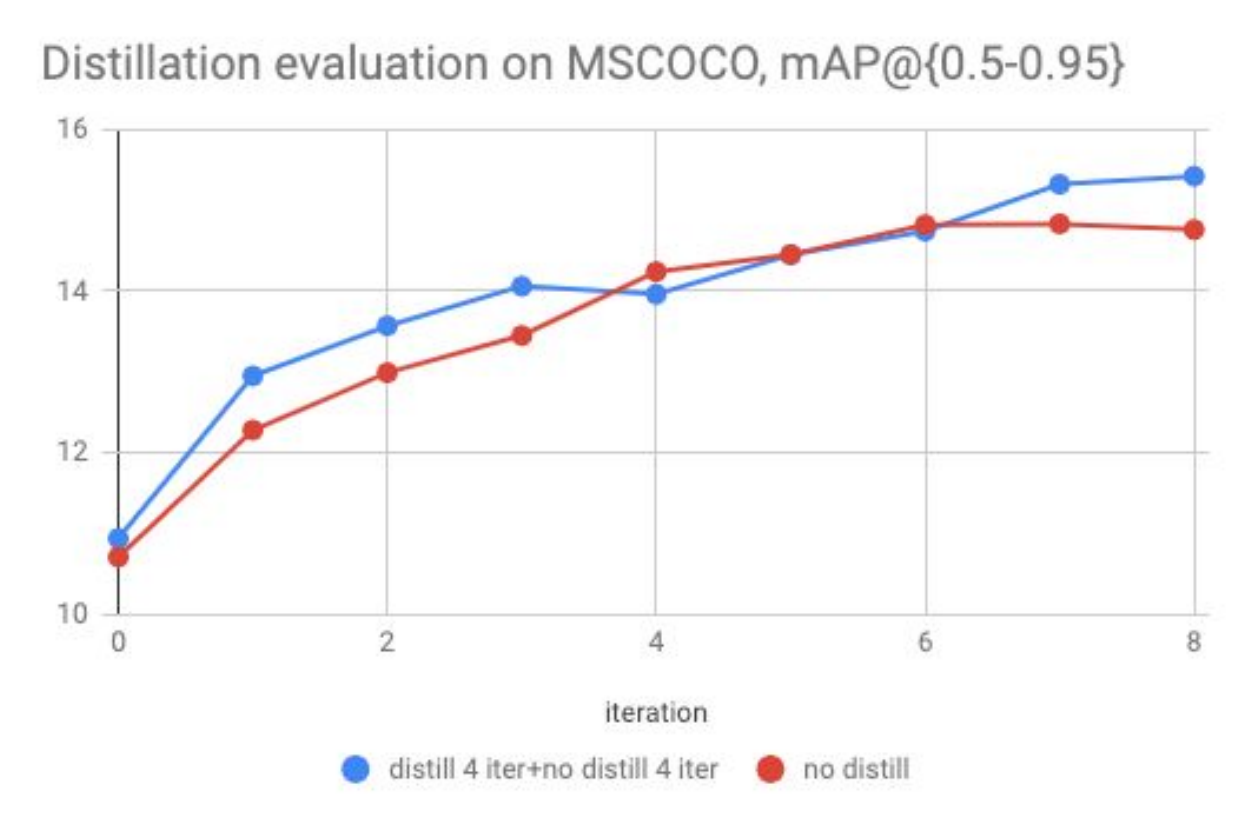}
    \caption{Comparison between ``half-distill" and ``no-distill" on target detector performance on MSCOCO 2017.}
      \label{fig:distill-coco}
\end{figure}

\textbf{Combining distillation in training-mining process.} 
We find that the quantity (\ie \# boxes) and quality (\ie box precision) of annotations are the two key factors that influence the detector performances: both higher quality and higher quantity result in better detectors. 
This inspires us to combine the distillation (higher quality) with non-distillation (larger quantity) method, called half-distill. 
We apply Det-AZ\&RPN-A\&Distill in the first 4 iterations and  Det-AZ\&RPN-A in the later 4 iterations. 
The experimental results are shown in Figure \ref{fig:distill-coco}. 
We can see that: 1) in the beginning stage (first three iterations), the performance of ``half-distill" is significantly better than that of ``no-distill", because ``half-distill" could generate higher quality of annotations; 
2) in the middle stage (around 4 iterations), ``no-distill" catches ``half-distill", as ``half-distill" suffers from fewer mined annotations; 3) in the final stage, after we switch the ``half-distill" to ``no-distill", the performance improves again. 

\textbf{Comparison with state-of-the-art methods.} 
The most related work is SemiMIL \cite{Uijlings_2018_CVPR}, but it doesn't use seed box annotations for the target categories. 
For a fair comparison, we build two stronger baseline methods based on SemiMIL \cite{Uijlings_2018_CVPR}. 1) SemiMIL+Seed+FRCN: We use SemiMIL to mine the box annotations from images, and then add the same seed annotations to the training set, following \cite{Uijlings_2018_CVPR} to train a standard Faster RCNN. 2) SemiMIL+Seed+NOTE-RCNN: Similar to the previous baseline, but we replace the standard Faster RCNN by NOTE-RCNN. 

\begin{table}[]
\begin{tabular}{l|l|l}
\hline
model                & backbone         & mAP  \\ \hline
LSDA \cite{hoffman2014lsda}                & alexnet          & 18.1 \\ \hline
Tang \etal\cite{tang2016large}  & alexnet          & 20.0 \\ \hline
FRCN+SemiMIL \cite{Uijlings_2018_CVPR}        & alexnet          & 23.3 \\ \hline
FRCN+SemiMIL  \cite{Uijlings_2018_CVPR}       & inception-resnet & 36.9 \\ \hline
FRCN+SemiMIL+Seed  & inception-resnet & 38.7 \\ \hline
NOTE-RCNN+SemiMIL+Seed & inception-resnet & 39.9 \\ \hline
Ours(wo/ distill)    & inception-resnet & 42.6 \\ \hline
Ours(w/distill)      & inception-resnet & 43.7 \\ \hline 
\end{tabular}
\\\caption{Comparison with state-of-the-art on ILSVRC 2013}
\label{tbl:sotaxx}
\end{table}

The performance of state-of-the-art methods and the new baselines are shown in Table \ref{tbl:sotaxx}. Comparing FRCN+SemiMIL+Seed and FRCN+SemiMIL, we can see that by adding seed annotations, the performance increases by 1.8\%. By changing Faster RCNN to NOTE-RCNN, the performance increases by 1.2\%, which shows the effectiveness of NOTE-RCNN in handling the noisy annotations. Our method (wo/ distill) achieves 42.6\% mAP and outperforms all state-of-the-art methods; by applying distillation (w/ distill), we further improve the performance to 43.7\%, which is the best among all methods.

\textbf{What is the limit of data scale?} 
Previous evaluation shows that our method can consistently improve performance when the number of seed bounding box annotations varies from 10 to 90. 
We also have an experiment to evaluate the effectiveness of the proposed method when a relatively large number of seed bounding is available.  
Specifically, we use the whole MSCOCO dataset (all 80 categories, around 15k boxes pe category) as the seed annotation set $\mathbf{B^0}$, and Openimage V4 as image-level annotation set $\mathbf{A}$, (only image-level labels is used). Det-AZ\&RPN-A is tested in this experiment, the results are shown in Table \ref{tbl:oid}. 
It can be seen that by our method can still consistently improve the detection performance during the training-mining process. 
However, the performance improvements saturates at iteration 2.
The reason is the initial detector trained with already have good accuracy,
it takes less iteration to mine enough useful bounding boxes.
It shows that even when a large amount of box-level annotations is available, 
larger scale of image-level annotation augmentation is still helpful for further performance improvement.

\begin{table}[]\small
\centering
\begin{tabular}{l|c|c|c|c}\hline
                 & mscoco & iter 1 & iter 2 & iter 3  \\ \hline
mAP@\{0.5-0.95\} &     32.2        &   33.6 & 34.0  & 34.0  \\\hline    
\end{tabular}
\vspace{0.5em}
\\\caption{``mscoco" means the Det-AZ\&RPN-A is only trained on MSCOCO, ``iter k" means iterative training of Det-AZ\&RPN-A on OpenImage image label set for k times. }
\label{tbl:oid}
\end{table}

\section{Conclusion and Future Work}
We proposed a new semi-supervised object detection formulation, which uses large number of image levels labels and a few seed box level annotations to train object detectors. 
%
%
To handle the label noises introduced in training-mining process, we proposed a NOTE-RCNN object detector architecture, which has three highlights: an ensemble of two classification heads and a distillation head to improve the mining precision, masking the negative sample loss in box predictor to avoid the harm of false negatives, and training box regression heads only on seed annotations to eliminate the harm from inaccurate box boundaries. Evaluations were done on ILSVRC and MSCOCO dataset, we showed the effectiveness of the proposed methods and achieved the state-of-the-art performance.
In the future, we plan to add human annotation to the training-mining iterations. We believe a combination of human annotation and accuracy box mining can further improve the detector performance.

{\small
\bibliographystyle{ieee}
\bibliography{egbib}
}

\end{document}